# A Pushing-Grasping Collaborative Method Based on Deep Q-Network Algorithm in Dual Perspectives

Peng Gang [a,b], Liao Jinhu [a,b], Guan Shangbin [a,b]

[a] Key Laboratory of Image Processing and Intelligent Control, Ministry of Education
[b] School of Artificial Intelligence and Automation, Huazhong University of Science and Technology, Wuhan, China

*Abstract*—**Aiming at the traditional grasping method for manipulators based on 2D camera, when faced with the scene of gathering or covering, it can hardly perform well in unstructured scenes that appear as gathering and covering, for the reason that can't recognize objects accurately in cluster scenes from a single perspective and the manipulators can't make the environment better for grasping. In this case, a novel method of pushing-grasping collaborative based on the deep Q-network in dual perspectives is proposed in this paper. This method adopts an improved deep Q-network algorithm, with an RGB-D camera to obtain the information of objects' RGB images and point cloud from two perspectives, and combines the pushing and grasping actions, so that the trained manipulator can make the scenes better for grasping, so that it can perform well in more complicated grasping scenes. What's more, we improved the reward function of the deep Q-network and propose the piecewise reward function to speed up the convergence of the deep Q-network. We trained different models and tried different methods in the V-REP simulation environment, and it drew a conclusion that the method proposed in this paper converges quickly and the success rate of grasping objects in unstructured scenes raises up to 83.5%. Besides, it shows the generalization ability and well performance when novel objects appear in the scenes that the manipulator has never grasped before.**

*Index Terms*— **Dual Perspectives, Deep Q-Network, Pushing-Grasping Collaborate, Reward Design, RGB-D Image.**

## I. INTRODUCTION

As one of the basic problems in the field of robot operation, grasping is the main means for robots to interact with the real world to complete operational tasks (such as sorting, assembly, etc.) [1]. The traditional 2D camera-based robot grasping methods perform well in grasping with high success rate when facing simple scenes. Kathleen A M et al. [2] uses thresholding and polygon encircling algorithms to locate the fruit to be grasped, and the grasping success rate reaches 85%. In [3], the improved HOG (Histogram of Oriented Gradient) is used to identify the industrial components for grasping, and the success rate of grasping components came up to 92%. In [4], a method is proposed aiming at grasping tomatoes, combining with two color spaces' images, using thresholding to locate and recognize the targets, and the success rate of grasping came up to 93%. However, in actual grasping applications, there are a large number of gathering and covering objects in grasping scene. Since 2D cameras cannot obtain information of height, it is difficult for robotic grasping system based on 2D cameras to grasp gathering and covering objects.

Deep reinforcement learning combines the perception ability of deep learning and the decision-making ability of reinforcement learning[5],[6], so that many grasping methods based on deep reinforcement learning emerged, and the depth camera can obtain deep information, which further improved the cognition level of the robot. In [7], a method of deep reinforcement learning in robotic grasping work was proposed for the first time, and it pointed out that the deep reinforcement learning algorithm model trained in a simulation environment can be applied to the grasping environment in reality through less training steps of grasping real objects. In [8], a method of knowledge-induced deep Q-network (DQN) was proposed. The network integrates perceptual information into DQN to provide prior knowledge for network training, so that the trained model has better generalization ability than traditional DQN, and the manipulator can better grasp unknown objects which do not exist in training. In [9], taking Actor-Critic in deep reinforcement learning as the main body, with redesigning the reward function, a reward function designing method based on objects structure matching was proposed. The improved algorithm converges faster with higher success rate of grasping. In [10], it improves the experience playback mechanism of the depth deterministic strategy gradient algorithm proposed by Google, which not only improves the performance of the algorithm, but also reduces the use of storage space and improves the efficiency of the manipulator to grasp objects. However, there are still problems in the current grasping methods based on deep reinforcement learning such as insufficient information utilization, network convergence slowly, and low success rate of grasping. In [11],[12], a method of using the affordance map to represent the grasp target by analyzing the whole scenes is proposed, which greatly improves the accuracy of the robot grabbing. Figure. However, because the environment is usually complex and unstructured, sometimes the grasping position indicated in the affordance map is difficult for manipulators to grasp. In [13], a method of active affordance exploration for robot grasping is proposed but it didn't provide the design guidelines of the reward. In[14], an active exploration algorithm which leverages the deep Q-Network (DQN) is proposed to facilitate the robot to actively explore the environment until a good affordance map is generated, but it just process images from a single perspective, so there are inevitably problems of information missing.

In order to solve the problems above, this paper proposes to obtain the top view of objects' point cloud from two perspectives, combine the pushing and grasping actions, use the piecewise reward function, and train the model of robot grasping iteratively. The trained model gives the best actions manipulator takes. The model we trained has a high success rate of grasping and has a certain generalization ability for grasping unknown objects.

## II. DESIGN OF PUSHING-GRASPING METHOD

When the robot is grasping objects in current scene, the state of scene after grasping depends on the current state of the scene and the actions taken by the manipulator. The grasping process can be described by Markov decision process (MDP) [15]. Markov decision process contains 5 elements: state set $S$, action set $A$, state transition probability matrix $P$, reward set $R$ and the discount factor $\gamma$. The grasping process of the manipulator can be described as in the state $s_t$. According to the strategy $\pi$, the manipulator takes action $a_t$ with probability $P(s_{t+1}|s_t, a_t)$ to act on the scene so that it changes into the new state $s_{t+1}$, and get the reward $r_t$. The goal of reinforcement learning for manipulators is to find an optimal strategy $G_t$ and the grasping action of manipulators can obtain the maximum return $G_t$ cumulatively, where $\gamma$ is a discount factor which aims to reduce the influences of current rewards acting on future behavior.

$$G_t = R_{t+1} + \gamma R_{t+2} + \gamma^2 R_{t+3} + \cdots = \sum_{k=0}^{T} \gamma^k R_{t+1+k} \quad (1)$$

According to Markov decision process, we can know that reinforcement learning needs to store the mapping relations between the state and the action, which is called the learning strategy. The mapping relations are generally stored in a table. For large-scale problems, the tables can hardly be maintained and stored in memory. Therefore, this paper adopts a method of deep reinforcement learning based on value function-deep Q network [16], [17]. DQN takes the feature vector $\emptyset(s)$ of state $s$ as input, and outputs the state-action value function $Q(s, a)$ which corresponds to each action through network processing.

DQN adopts an approximate description of the value function, so it evolves into Regression problem, called function fitting:

$$Q(s, a; \theta) \approx Q^*(s, a) \quad (2)$$

Where $\theta$ represents model parameters. In order to speed up convergence of the network, we designed two convolutional neural networks: current value network function $Q(s, a; \theta)$ and target value network function $Q(s, a; \theta^-)$. The parameters of the current value network will be updated with the iteration process, and the target value network will calculate the $Q$ value along the iteration process, but the network parameters will be fixed to a certain number of iteration steps. After certain steps, the parameters of the current value network are copied and updated, so the optimization goal of model is:

$$C = R + \gamma max_{a'} Q(s', a'; \theta^-) \quad (3)$$

Therefore, by minimizing the square error of the current value function and the target value function, the network parameters can be updated.

$$L(\theta) = E\left[\left(R + \gamma max_{a'} Q(s', a'; \theta^-) - Q(s, a; \theta)\right)^2\right] \quad (4)$$

### A. Description of State Set

In this paper, we use RGB-D cameras to obtain the point cloud of objects, but the point cloud image cannot be put directly into the DQN as input, so this paper takes the RGB top view of point cloud in current scene before each action as a state $s_t$ as the state input of the DQN. But for gathering objects, if the camera is still above the grasping area and installed vertically downwards, it will make a difference to plans of grasping. Therefore, a method with certain angle installation which is placed outside the grasping area is adopted, but there will be information missing when the top view is obtained under a single angle view, as shown in Figure 1. This paper uses two RGB-D cameras to capture from two perspectives so that complete object information is guaranteed. The pixel resolution of the RGB top view obtained is 224×224.

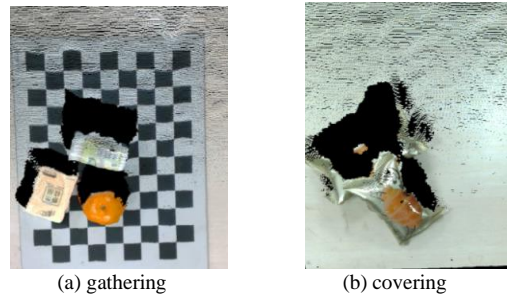

(a) gathering        (b) covering

Fig. 1. Information Missing in Single Perspective

### B. Description of Action Space

For the purpose of adapting complex grasping scenes, this paper introduces a manipulator pushing based on the grasping. For the reasons that pushing can disrupt the original arrangement of the objects, which makes it more convenient for manipulator to grasp. Two pieces of information need to be obtained when the manipulator takes action of grasping or pushing, namely target position $(x, y, z)$ that the gripper needs to move to and the postures of the gripper. The target position $(x, y)$ can be learned by DQN, and the height $z$ can be obtained from depth maps; for the postures of the end manipulator, this paper adopts a vertical downward grasping method, so there is only a rotation angle $\theta$ along z-axis in the postures of the end manipulator, and the manipulator can push along the direction of an angle $\alpha$ while it is pushing.

When the manipulator is pushing or grasping objects, it is not necessary for the end rotation angle and the pushing angle of the manipulator to be an accurate angle. Therefore, we discretize the angles above and divides $360°$ into 16 equal parts, which are defined by $22.5°$ as a grasping or pushing direction. Therefore, the action space can be described as:

Pushing action: taking coordinates $(x, y)$ learned by the DQN as the starting position, the end of the manipulator pushes the object in one of the 16 directions set, and the pushing distance is half the length of the longest side of the objects, which is set to 5cm in this paper;

Grasping action: Taking coordinates $(x, y)$ learned by the DQN as the target position of the center movement of the gripper at the end of the manipulator, which is rotated to one of the 16 directions above to grasp the objects.

It can be seen from the state set that the descriptions of state use a top view, while the visual calibration uses an oblique view as usual. While manipulator taking actions to push or grasp, we need to determine the conversion between the two images, and the conversion relationship can be determined by perspective transformation.

### C. Wheel Speed Anomaly Detection

Due to the recognition of images, in this paper, we model the Q function as two Fully Convolutional Networks (FCN) [18], one of which is used to select the pushing action and the other is used to select the grasping action. This two networks have the same structure: each network contains a pre-trained DenseNet121[19], and then is channel-cascaded with two $1 \times 1$ convolutional layers, and each convolutional layer has a ReLU activation function and batch normalization, and finally bilinear upsampling, each network will capture the RGB top view of point cloud of the first two perspective scenes as input.

In order to obtain the suitable angle of pushing or grasping, we rotate the RGB top view 15 times as the input of FCN. That is, for each FCN network, $16 \times 2$ pictures will be set as input, and output will be the same size and number of heat maps as input images. Each pixel of pictures represents an independent Q value prediction. Sorting the Q values of the pixels in the output heat maps of the two networks, we can respectively obtain the heat maps' largest Q value in the pushing network and grasping network, and we can select the Q value in the two heat maps who is larger as the final network output (Figure 2.). The final selected pixel and direction are given by the pixel corresponding to the maximum Q value in the heat maps which the network finally output, and the rotation angle corresponding to the heat maps. The selected action is given by the network corresponding to the heat map.

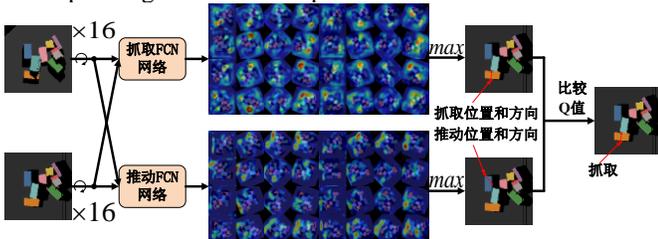

Fig.2. Process of the DQN Algorithm

### D. Reward Configuration

The traditional reward is set as a reward $R = 1$ for a successful grasping. If the total change of the difference between height maps exceeds a certain threshold $\tau$, a reward $R = 0.5$ is given. Such a reward is too sparse, which will lead to non-convergence and slow convergence in model training. In order to deal with the problems, a piecewise reward strategy instead of a single reward is proposed in this paper. If grasp successfully, the reward $R = 1$ is given, if grasp failed, the reward $R = -1$ is given; if the rate of pixel change $\tau$ of the scene after pushing is 10%~24%, the reward $R = 0.3$ is given; If the rate of pixel change $\tau$ is between 24%~40%, the reward $R = 0.5$ is given; if the rate of change $\tau$ is 40%~100%, the reward $R = 0.7$ is given; in other cases, the reward $R = -0.1$ is given.

### III. Configuration of Simulation Environment and Network Training

In order to verify that the DQN algorithm after adding pushing action and dual perspectives has a higher success rate of grasping and generalization ability, a comparative experiment of the two algorithms is carried out on the simulation platform for the grasping in the covering scene. The hardware used is the Intel i7-8700k as CPU and Nvidia GTX 1080Ti 16GB as GPU. V-REP[20] is used for training and verification under the environment of Ubuntu 16.04. As shown in Figure 3, the methods above are used to learn how to grasp objects in the area.

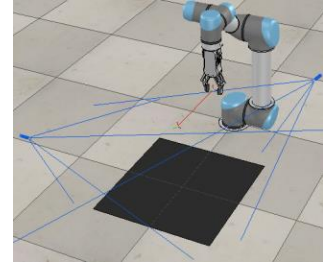

Fig. 3. Simulation Environment of Grasping

Stochastic gradient descent is used to train the FCN network during training. In the experiment, the learning rate is set to $10^{-4}$, and the weight decay is set to $2^{-5}$. The exploration strategy adopts the $\epsilon$ −greedy algorithm. The initial $\epsilon$ is set to 0.65, the annealing is set to 0.1 after training, and the discount factor $\gamma$ is set to 0.6.

### IV. Results and Analysis of Experiment in Simulation Environment

Compared with the grasping only strategy, the DQN algorithm based on dual perspectives in this paper improves in 3 aspects: ①The dual perspectives are used to obtain the objects' information in the area to be grasped, avoiding missing information in single perspective; ② Introducing a novel pushing action, avoiding the limitations of grasping only; ③ Adopt multi-level reward strategy, avoiding the problems of slow convergence caused by single reward. In order to verify the influence of the improvement above on the model of grasping performance, this section designs a set of comparative experiment for verification.

### A. Comparative Experiment

In this experiment, the pushing and grasping strategy based on DQN and dual perspectives was compared with three methods of removing the improvement aspects to verify the influence of each improvement aspect on the performance of the algorithm. The three methods to remove from the improvement aspects are the and pushing-grasping strategy based on DQN and single perspective, the grasping-only strategy based on DQN and dual perspectives, and the pushing-grasping strategy based on DQN and dual perspectives but with only a single reward. The training scene is set as follows: select 10 objects randomly arranged and put them in the grasping area. Manipulator learns how to grasp the objects in the area through trial and error, until the objects in the work space are grasped completely or out of the view, and then select 10 objects randomly again and place them in the grasping area. Every 100 objects grasped is a group of training. After one group of training, the next group of training starts. The training process is shown in Figure 4.

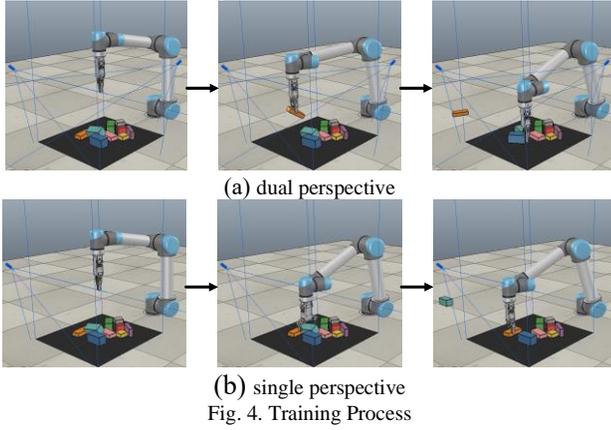

(a) dual perspective

(b) single perspective

Fig. 4. Training Process

The evaluation of performance is mainly to evaluate the success rate of grasping and the convergence speed of the model. The success rate of grasping mainly refers to the ratio of the number of objects grasped successfully to the number of objects to be grasped. The model training results of the four methods are as follows Shown in Figure 5.

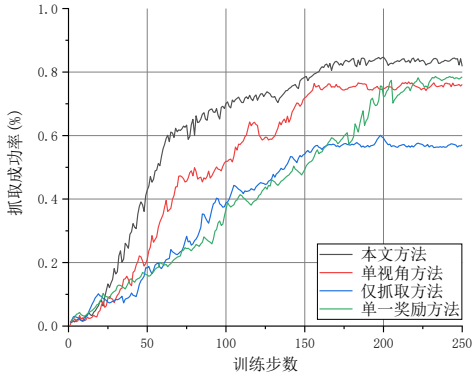

Fig. 5. Success Rate of Different Methods to Grasp

We can find out that compared with the single perspective and grasping-only strategies, the pushing-grasping strategies based on DQN and dual perspective have higher success rate of grasping. The single-perspective strategy achieved a success rate of 76.1% in grasping after 170 iterations. The grasping-only strategy achieved a success rate of 57.3% in grasping after 180 iterations. However, the grasping strategy based on DQN and dual perspectives achieved a success rate of 83.5% in grasping after 180 iterations. The results show that the introduction of the push operation has a great contribution to the success rate of grasping. For pushing actions disrupting the original structure of the objects' arrangement, it makes the sense better for grasping objects. The dual perspectives are also beneficial to improving the success rate of grasping, which is probably for the reasons that dual perspectives can obtain more comprehensive information of objects than single perspective.

Compared with the strategy of single reward, the pushing-grasping strategy based on DQN and dual perspectives has a faster convergence speed, but it should be noted that the success rate of grasping in the single reward strategy doesn't drop too much, which is also stable at 78.5%. The results show that the multi-level reward can make the model have a faster convergence speed.

### B. Tests of Model's Generalization Ability

After the model is trained, in order to verify that the pushing-grasping strategy based on DQN and dual perspectives has a better generalization ability than the grasping strategy based on DQN and single perspective, three unknown objects are selected to design three groups of experiments, and three unknown objects which are triangles, semicircles, and cylinders. The settings of scene for the three experiments are as follows: ①1 unknown object + 5 known objects placed in the working space; ②3 unknown objects + 3 known objects placed in the working space; ③5 unknown objects + 1 known object placed in the working space. Use trained model above to make the manipulator to grasp objects in the workspace until the objects in the workspace are grasped successfully or the objects are out of the workspace. In each group of the experiments, each unknown object needs to appear 30 times in total for the convenience of statistics. The experiment in the simulation environment is shown in Figure 6.

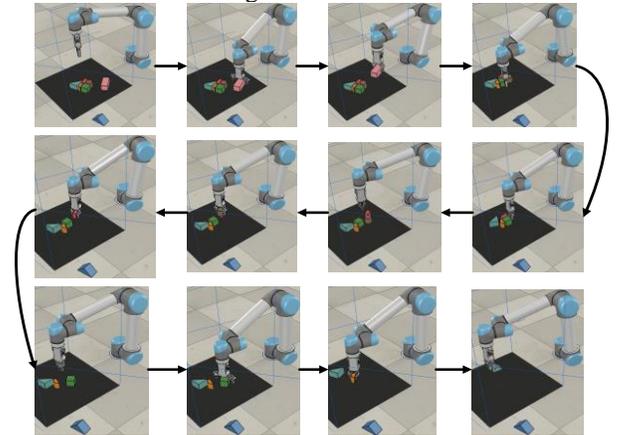

Fig.6. Grasping of Unknown Objects

The results of the success rate of grasping with the two methods are shown in Table 1.

TABLE I
SUCCESS RATE OF GRASPING UNKNOWN OBJECTS
(U: UNKNOWN OBJECTS, K: KNOWN OBJECTS EXIST IN TRAINING MODEL)

| Test group | Triangle | Semicircle | Cylinder |
|---|---|---|---|
| 1U+5K(Our Method) | 11/30 | 14/30 | 17/30 |
| 1U+5K(Grasp Only) | 3/30 | 7/30 | 6/30 |
| 3U+3K(Our Method) | 6/30 | 10/30 | 14/30 |
| 3U+3K(Grasp Only) | 2/30 | 4/30 | 6/30 |
| 5U+1K (Our Method) | 5/30 | 6/30 | 12/30 |
| 5U+1K (Grasp Only) | 1/30 | 2/30 | 4/30 |

From data in table I, it's obvious that the DQN-based grasping strategy has a low success rate of grasping when grasping unknown objects, while the pushing-grasping strategy based on DQN and dual perspectives has a better generalization ability when grasping unknown objects, with a higher success rate especially in grasping cylinders which are similar to the objects in training set, but it should be noted that when there are an increasing number of unknown objects, the success rate of grasping unknown objects is also getting lower. It's probably for the reason that the arrangement structure of unknown objects is different from the arrangement structure of the objects in the training set.

## V. CONCLUSION

In order to solve the problems of insufficient information utilization, slow network convergence and low success rate of grasping in current grasping strategy based on deep reinforcement learning, a pushing-grasping collaborative sorting method based on DQN and dual perspectives is proposed in this paper. This strategy obtains the top views of the objects' point cloud from two perspectives, combining pushing and grasping actions, with a piecewise reward function, and finally trains to obtain a robot grasping model iteratively, which gives the best action taken by manipulator. Through comparative experiments in V-REP simulation environment, it can be seen from the results that the method proposed in this paper can not only speed up the convergence of the network, but also disrupt the original arrangement structure of the objects by pushing actions, which improves the success rate of grasping obviously, which can come up to 83.5%. And it also has a certain generalization ability when grasping unknown objects.

In future work, in order to further improve the cognitive ability of the manipulator, a target detection framework can be introduced, so that the manipulator can also recognize the types of objects to be grasped, which is more convenient for objects' classification.

**Author Biographies**

**Peng Gang** (1973-) received the doctoral degree from the Department of control science and engineering of Huazhong University of Science and Technology (HUST) in 2002. Currently, he is an associate professor in the Department of Automatic Control, School of Artificial Intelligence and Automation, HUST. He is also a senior member of the China Embedded System Industry Alliance and the China Software Industry Embedded System Association, a senior member of the Chinese Electronics Association, and a member of the Intelligent Robot Professional Committee of Chinese Association for Artificial Intelligence. His research interests include intelligent robots, machine vision, multi-sensor fusion, machine learning and artificial intelligence.

**Liao Jinhu** (1994-) received his bachelor degree in school of automation from Center South University, China, in 2017. He received master's degree from the Department of Automatic Control, School of Artificial Intelligence and Automation, Huazhong University of Science and Technology. His research interests include intelligent robots and perception algorithms.

**Guan Shangbin** (1997-) received his bachelor degree in school of automation from Wuhan University of Science and Technology, Wuhan, China, in 2020. He is currently a graduate student at the Department of Automatic Control, School of Artificial Intelligence and Automation, Huazhong University of Science and Technology, Wuhan, China. His research interests include intelligent robots and perception algorithms.